\documentclass[
twocolumn,
]{ceurart}

\usepackage{listings}
\usepackage[utf8]{inputenc} 
\usepackage[T1]{fontenc}    
\usepackage{hyperref}       
\usepackage{url}            
\usepackage{booktabs}       
\usepackage{amsfonts}       
\usepackage{nicefrac}       
\usepackage{microtype}      
\usepackage[ruled,vlined,linesnumbered]{algorithm2e}

\begin{document}

\copyrightyear{2022}
\copyrightclause{Copyright for this paper by its authors.
  Use permitted under Creative Commons License Attribution 4.0
  International (CC BY 4.0).}

\conference{4th Edition of Knowledge-aware and Conversational Recommender Systems (KaRS) Workshop @ RecSys 2022, September 18--23 2023, Seattle, WA, USA.}

\title{A Nonparametric Contextual Bandit with Arm-level Eligibility Control for Customer Service Routing}

\author[1]{Ruofeng Wen}[%
email=ruofeng@amazon.com,
]
\cormark[1]
\address[1]{Customer Engagement Technology, Amazon}

\author[1]{Wenjun Zeng}[%
email=zengwenj@amazon.com,
]

\author[1]{Yi Liu}[%
email=yiam@amazon.com
]

\cortext[1]{Corresponding author.}

\begin{abstract} 
Amazon Customer Service (CS) provides real-time support for millions of customer contacts every year. While bot-resolver helps automate some traffic, we still see high demand for human agents, also called subject matter experts (SMEs). Customers outreach with questions in different domains (return policy, device troubleshooting, etc.). Depending on their training, not all SMEs are eligible to handle all contacts. Routing contacts to eligible SMEs turns out to be a non-trivial problem because SMEs' domain eligibility is subject to training quality and can change over time. To optimally recommend SMEs while simultaneously learning the true eligibility status, we propose to formulate the routing problem with a nonparametric contextual bandit algorithm (K-Boot) plus an eligibility control (EC) algorithm. K-Boot models reward with a kernel smoother on similar past samples selected by $k$-NN, and Bootstrap Thompson Sampling for exploration. EC filters arms (SMEs) by the initially system-claimed eligibility and dynamically validates the reliability of this information. The proposed K-Boot is a general bandit algorithm, and EC is applicable to other bandits. Our simulation studies show that K-Boot performs on par with state-of-the-art Bandit models, and EC boosts K-Boot performance when stochastic eligibility signal exists.

\end{abstract}


\maketitle

\hypertarget{introduction}{%
\section{Introduction}\label{introduction}}

In Amazon Customer Service (CS), we dispatch human agents, also called subject matter experts (SMEs) in real-time to handle millions of customer contacts. The SME routing automation process has two steps: first there is a natural language understanding (NLU) model to process customer's input and identify the relevant domain (return policy, device troubleshooting, etc.); then it dispatches an SME who is eligible. We define \emph{eligibility} when the SME masters the required skill in the relevant domain through training. SME routing turns out to be a non-trivial problem for four reasons. First, the NLU model is unlikely to categorize the domain with perfect accuracy. Second, the reliability in SME eligibility as identified by operation team is subject to training program quality and readiness. Third, the domain taxonomy and SMEs' eligibility status can change in a decentralized way. In reality, it is difficult to keep track of all the eligibility updates, assuming correctness. Finally, eligible SMEs do not guarantee customer satisfaction, leading to noisy feedback signals. All these uncertainties make SME routing a challenging problem.

\begin{figure}
     \centering
     \includegraphics[width=0.5\textwidth]{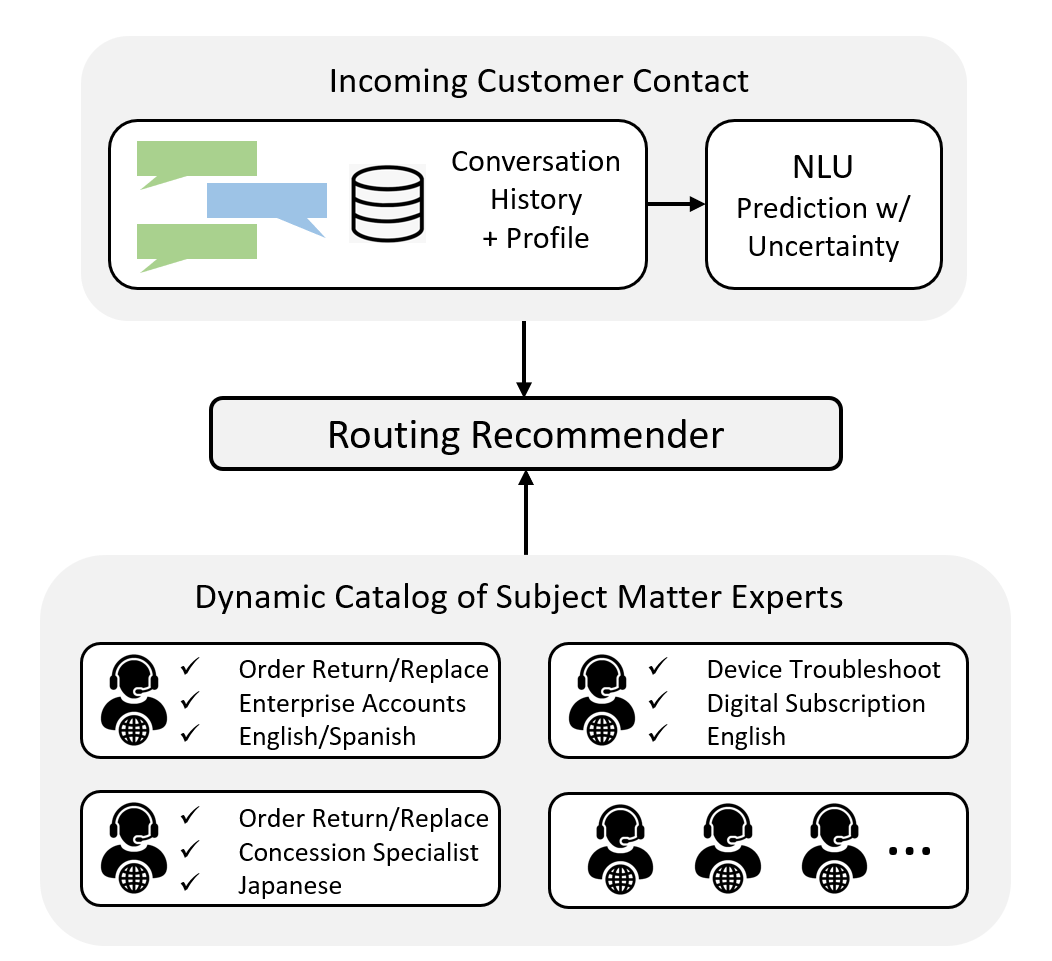}
     \caption{The CS Routing Problem. The goal of the recommender is to select an agent that will result in the best outcome of this customer contact.}
     \label{fig:figure1}
\end{figure}

To tackle the complexity, we formulate CS routing as a recommendation problem and use contextual Bandit as a solution. Bandit, a framework for sequential decision making, has been used for online recommendation across companies such as Amazon, Google, Netflix and Yahoo~\cite{elena2021survey}. In contextual bandit, the decision maker sequentially chooses an arm/action (SME in our case), based on available contextual information (contact transcript embedding, customer profile, etc.), and observes a reward signal (customer satisfaction, contact transfer, etc.)~\cite{liu2021map}. The objective is to recommend arms at each step to maximize the expected cumulative reward over time. We apply bandit to the routing problem because it can optimally explore uncertainties and adapt to dynamic changes~\cite{lattimore2020bandit}. Particularly, the uncertainties in SME eligibility motivate us to formulate a new type of bandit to utilize arm-level eligibility signals.

\textbf{Our contribution}. We propose K-Boot, a nonparametric contextual bandit algorithm to model reward and explore, and an Eligibility Controller (EC) algorithm to model arm-level eligibility. K-Boot uses a $k$-nearest neighbors ($k$-NN) approach to find similar samples from the past, applies a Nadaraya–Watson kernel regression among the found samples to estimate the reward, and adopts Bootstrap Thompson Sampling as the exploration strategy. The EC component implements a dynamic top arm filter based on its estimated Spearman's Rank Correlation Coefficient between eligibility and reward. We are interested in this end-to-end nonparametric setup for practicality: robustness in performance, strong interpretability and simple non-technical maintenance which is friendly to business partners. For instance, the $k$-NN component makes it easy to investigate which historical contacts support the SME recommendation, and then deploy instant online fixes by trimming unwanted outliers. While K-Boot and EC are proposed here as a suite, each can be applied independently - K-Boot is a general Bandit algorithm and EC can control arm-eligibility for other bandits, such as LinUCB~\cite{li2010contextual}. 

In the remainder of the paper, we review related literature in Section \ref{related-work-yiam}, set up the formal problem in Section \ref{formulation-contextual-bandit-with-arm-eligibility} and detail the proposed algorithms in Section \ref{method}. We then present our model results in Section \ref{sec:Experiments}, and conclude the paper last.

\hypertarget{related-work-yiam}{%
\section{Related Work}\label{related-work-yiam}}

In most bandit applications, the items to recommend are products and webpage content (widget, news articles, etc.). Recently, we see research efforts in using bandit for routing recommendation in a \emph{conversation}~\cite{sajeev2021contextual, upadhyay2019bandit}. In \cite{upadhyay2019bandit}, they used contextual bandit to utilize query and user features along with the conversational context and build an orchestration to route the dialog. In \cite{sajeev2021contextual}, they built bandit models to assign solutions given customer's queries on Microsoft's product to a chatbot. This is the most relevant work to our problem. However, they assume an upstream model to provide a set of plausible actions to start with and the bandit itself does not deal with the uncertainty in arm eligibility. Overall, comparing to the extensive bandit literature existing for product and webpage content recommendation, few bandit works are there for routing recommendation.

Thompson Sampling (TS) is a heuristic for balancing exploration and exploitation in bandits~\cite{chapelle2011empirical}. In the parametric manner, TS is implemented by drawing the reward function parameter \(\theta\) from a posterior distribution, calculating rewards with the sampled parameters, and selecting the arm that maximizes the estimated reward function. \(\theta\) is often assumed to follow Conjugate-Exponential Family distributions resulting closed-form posteriors. There are also work where variational methods are used to provide an analytical approximation to the posterior and enforce trackability~\cite{graepel2010blip}. Bootstrap TS introduces volatility in a different way: it pulls the arm with the highest bootstrap mean, estimated from reward history in a nonparametric way ~\cite{osband2015bootstrapped}. One drawback of the approach is the computational cost to train one bandit model per Bootstrapped training set, which is difficult for large-scale real-time problems. \cite{kveton2019garbage} proposed a general nonparametric bandit algorithm but they fell back to the parametric reward function approach when there is context. \cite{riou2020bandit} further improved the exploration efficiency with Bayesian Bootstrap, but also only for non-contextual bandit. We pair Bootstrap TS with a single $k$-NN for nonparametric contextual exploration.

Kernel-based Bandit uses a nonparametric reward estimator via kernel functions, combined with an exploration policy like TS or UCB. Gaussian Process (GP) has wide applications in bandit domain, e.g.~GP-UCB~\cite{srinivas2009gaussian} and GP-TS~\cite{chowdhury2017kernelized}. However, GP inference is expensive - the standard approach has \(O(N^3)\) complexity, where \(N\) is the number of training samples. The most efficient sparse or low-rank GP approach still requires \(O(ND^2)\), where \(D\) is a hyperparameter indicating the number of \emph{inducing variables} - see \cite{hensman2013gaussian} for a review. We use $k$-NN to enable \(O(k\log N)\) inference time.

\hypertarget{formulation-contextual-bandit-with-arm-eligibility}{%
\section{Formulation: Contextual Bandit with Arm
Eligibility}\label{formulation-contextual-bandit-with-arm-eligibility}}

We define two types of eligibility setup with different data interfaces. \textbf{(1) Eligibility Scores}: for a given context $\mathbf{x}$, a list of eligibility scores for each arm \(\mathbf{e} = (e_1,e_2,\dots)\) is observed, where \(e_j \in [0,1]\) is for arm \(a_j\). A larger score means higher confidence in the arm being eligible for the current sample/iteration. An intuitive interpretation is that the arms are bidding for themselves. This score setup assumes (the owner of) each arm is able to adaptively assess the dynamic environment in a distributed and independent manner that can be unknown to the bandit. Typical examples include ads bidding and voice assistant's skills detection - each ad/skill may be managed via the same API, but by different clients under dynamic competition. \textbf{(2) Eligibility States}: assume the system at any time can be classified into one of \(L\) possible states. Each arm \emph{claims} whether it is eligible under a state, represented by a constant binary vector \(\mathbf{c}_j\) of length \(L\) for arm \(a_j\). If the \(l\)-th element of \(\mathbf{c}_j\) is \(1\), arm \(a_j\) claims to be eligible under the \(l\)-th state, and \(0\) otherwise. Given a context $\mathbf{x}$, a stochastic distribution over the \(L\) states is observed, represented by a probability vector \(\mathbf{p}\) of length \(L\). For CS routing, the state is the type of customer issue, the claim is SME's corresponding skill-set, and $\mathbf{p}$ is the NLU predictive distribution. Note that a score can be computed as the inner product of the claimed eligibility binary vector and state probability vectors: \(e_j := \mathbf{c}_{j}^{\intercal}\mathbf{p}\), so the state setup converges to the score setup. The major practical difference between the two interfaces is the amount of effort in the claiming side: the state interface simplifies the eligibility claims to static binary votes on a finite set of states, while the score interface mandates scoring with context awareness.

Consider a contextual bandit with a set of arms \(\{a_1,a_2,\dots\}\). At the $n$th round, context \(\mathbf{x}_n\) is observed and eligibility score \(\mathbf{e}_n\) is either observed (score interface) or derived (state interface). After an arm is pulled, a reward \(r_n \in [0,1]\) is revealed. The goal is to minimize the expected regret over $N$ rounds: $\sum_{n=1}^{N} (\mathop{\mathbb{E}}[r|a^{n}_*,\mathbf{x}_n,\mathbf{e}_n]-\mathop{\mathbb{E}}[r|a^n,\mathbf{x}_n,\mathbf{e}_n])$, where $a^{n}_*$ is the optimal action for this round in hindsight.

\hypertarget{method}{%
\section{Methodology}\label{method}}

The proposed algorithms run in order: Eligibility Controller (EC) to filter out ineligible arms, and K-Boot to find the optimal one within the rest. Below, we first introduce K-Boot to set the base and then EC.

\hypertarget{k-boot}{%
\subsection{K-Boot}\label{k-boot}}

K-Boot is a nonparametric bandit model. For a given context $\mathbf{x}$, it estimates the reward for arm \(a\) as the Nadaraya-Watson kernel weighted average over the observed rewards of the \(k\) nearest neighbors of $\mathbf{x}$ from all historical samples where arm \(a\) was pulled. We use Bootstrap TS as the exploration strategy. \(k\)-NN regression was known to fit well with Bootstrap-Aggregation, with an analytical form of smoothed conditional mean estimate that does not require actual resampling~\cite{steele2009exact, biau2010rate}. However, sampling from the confidence distribution of the mean estimate has little discussion. We devise a trick to shrink the resampling range from all historical samples to a local wrapper around the $k$-NN.

K-Boot is detailed in Algorithm 1. At iteration \(n\) with context $\mathbf{x}_n$, for the $m$th arm \(a_m\), let $\mathcal{D}_m$ be the set of historical samples where $a_m$ was pulled, and \(N_m = |\mathcal{D}_m|\). If \(N_m\) is zero, we sample reward from a standard uniform distribution (Line 6). If \(N_m\) is greater than \(K\), we first build an \textit{influential} subset $\mathcal{D}_m^\prime$ containing the $K^\prime$-NN of $\mathbf{x}_n$ (\(K < K^\prime\)). Intuitively, the $K^\prime$-NN serves as a buffer to cover enough sample variance around the $K$-NN, so that a Bootstrap on the influential samples is a good approximate of that on all samples. Formally, a sample is considered influential to make inference about $\mathbf{x}_n$, if its probability of being in the \(K\)-NN of $\mathbf{x}_n$ \emph{after} a Bootstrap on $\mathcal{D}_m$ is greater than \(\varepsilon\). This probability is computed analytically on Line 9, based on the derived equation (6) in \cite{steele2009exact}. Here the tolerance hyperparameter $\varepsilon$ controls the risk of missing influential samples and thus the fidelity of approximated Bootstrap TS. In our experiments with $N_m \leq 10^4$ and $\varepsilon = 0.01$, $K^\prime$ is empirically well bounded by $2K$. This process shrinks all later computation to $K^\prime$ samples, with the overhead neighbors search cost $O(\log N_m)$ per sample (we used Hnswlib \cite{malkov2018efficient} for approximated search). If \(N_m\) is less than \(K\), $\mathcal{D}_m$ itself is the influential set. We then add two pseudo samples to $\mathcal{D}_m^\prime$ on Line 14-15, in order to expand reward's empirical range for Bootstrap exploration~\cite{kveton2019garbage}. To give pseudo-samples proper contexts thus kernel weights, we set their distance to $\mathbf{x}$ as that of a random observation in $\mathcal{D}_m^\prime$, so its weight shrinks as more data seen. On Line 16-18, we then draw a Bootstrap resample and select $K$ nearest samples from it to calculate the estimated reward for \(a_m\) as the kernel weighted average. we implement $K_h(\cdot,\cdot)$ as the simplest \(O(K)\) Nadaraya-Watson kernel estimator with automatic bandwidth selection~\cite{silverman2018density} - more advanced models can be plugged in.

As we model reward for each arm separately as a common practice, K-Boot is flexible to add and removal of arms. This benefits applications like ours with decentralized and independent arms management. In addition, K-Boot has simple maintenance: (1) the nonparametric model has minimal assumption in data and only a single important hyperparameter \(K\) to balance between accuracy and computation. \(\varepsilon = 0.01\) works well across our experiments. (2) the algorithm is business friendly in terms of decision interpretability by neighboring examples and allowing intuitive online instant modification of model behaviors. For example, CS business owners may hide contacts with undesired trending patterns during a specific period (e.g. under a legacy policy), or let newly available SMEs "inherit" past example contacts or eligibility claims (next section) via domain knowledge (e.g. a rehired agent or additional training programs). The changes on data visibility will instantly and predictably affect the bandit model in production, with clear attribution. This data-driven while human-in-the-loop fungibility is essential to fast-paced operational business like customer service. 

\begin{algorithm*}
\DontPrintSemicolon
\SetAlgoLined
\SetKwInOut{Input}{Input}
\caption{K-Boot: a fully nonparametric contextual bandit}
\Input{Number of iterations $N$, number of arms $M$, number of nearest neighbors $K$, kernel function with bandwidth $K_h(\cdot,\cdot)$, regularized incomplete beta function $F_{\alpha,\beta}(\cdot)$, approximation tolerance $\varepsilon$.} 
\BlankLine
Initialize sample pool $\mathcal{D}_m := \varnothing$ and its size $N_m := 0$, for each arm $m = 1,\dots,M$\;
\For{$n = 1,\cdots,N$}{
    Observe context $\mathbf{x}_n$\;
    \For{$m = 1,\cdots,M$}{
        \uIf{$N_m = 0$}{
            Estimate reward: $\hat{r}_{m,n} \sim \mathcal{U}(0,1)$\;
        }
        \Else{
            \uIf{$N_m > K$}{
                $K^\prime := \min(k^\prime)$, s.t. $\sum_{i=1}^K{[F_{i,N_m-i+1}(\frac{k^\prime}{N_m}) - F_{i,N_m-i+1}(\frac{k^\prime - 1}{N_m})]} > 1 - \varepsilon$\;
                Find influential neighbors $\mathcal{D}_{m}^\prime :=$ the top $K^\prime$ samples in $\mathcal{D}_m$, with the largest $K_h(\mathbf{x}_n,\cdot)$\;
            }
            \Else{
                Set $\mathcal{D}_m^{\prime} := \mathcal{D}_m$\;
            }
            Draw a random sample $(\mathbf{x}_\star,r_\star)$ from $\mathcal{D}_{m}^\prime$\;
            Add pseudo-samples: $\mathcal{D}_m^{\prime} := \mathcal{D}_m^{\prime} \cup \{(\mathbf{x}_\star,0),(\mathbf{x}_\star,1)\}$\;
            Draw a Bootstrap sample $\mathcal{D}_m^{\star}$ from $\mathcal{D}_m^{\prime}$\;
            Find neighbors $\mathcal{D}_{m,K}^{\star} :=$ the top $\min(K,|\mathcal{D}_m^{\star}|)$ samples in $\mathcal{D}_m^{\star}$, with the largest $K_h(\mathbf{x}_n,\cdot)$\;
            Estimate reward: $\hat{r}_{m,n} := \sum_i{r_i K_h(\mathbf{x}_n,\mathbf{x}_i)}/\sum_i{K_h(\mathbf{x}_n,\mathbf{x}_i)}$, summing over $\mathcal{D}_{m,K}^{\star}$\;
        }
    }
    Pull arm $m^\star := \arg\max_m \hat{r}_{m,n}$, and observe true reward $r_n \in [0,1]$\;
    Update $\mathcal{D}_{m^\star} := \mathcal{D}_{m^\star} \cup \{(\mathbf{x}_n,r_n)\}$ and $N_{m^\star} := N_{m^\star} + 1$\;
}
\end{algorithm*}

\hypertarget{eligibility-control}{%
\subsection{Eligibility Control}\label{eligibility-control}}

EC is used to recognize arm-eligibility and the associated uncertainties, for optimized bandit exploration. In an ordinary contextual bandit model, the eligibility information can be trivially considered as part of the context \(\mathbf{x}\) - it is assumed to be positively correlated with \(r\), thus predictive as a plain input to a reward estimator. This assumes the reward model is able to eventually learn the relation between \(r\) and \(e\) after enough rounds. This trivial approach does not directly utilize eligibility to limit the range of arm exploration, and may unnecessarily explore inappropriate actions with catastrophic regret during cold-starts. The other extreme is to strictly follow the eligibility information and ignore the reward feedback: for the eligibility score interface, this could be simply pulling the arm with the highest score as if it were an oracle; for the eligibility state interface, this could be finding the state with the highest probability then remove arms that did not claim that state. Therefore, the empirical data would never be used to validate and correct the potentially biased business logic - a common pitfall in practice.

The main idea of EC is to leave only the top-\(k\) arms with the highest eligibility scores before applying a normal bandit, and adjust \(k\) periodically based on the empirical correlation between eligibility scores and rewards. We use Spearman's Rank Correlation Coefficient, denoted as \(\rho\). Using top-\(k\) arms is a heuristic trade-off between \(k = M\) (the trivial case; \(M\) is the total number of arms), and \(k = 1\) (the strict rule case). Intuitively, for the case of perfect correlation between score and reward (\(\rho = 1\)), \(k = 1\) is the oracle solution. If eligibility score has zero or even negative correlation (\(\rho \leq 0\)) with reward, \(k = M\) is the optimal solution - otherwise the arm exploration is restricted adverserially or randomly for no benefits. In the no-correlation case, the probability of ``the best arm by reward is not in the top-\(k\) by score'', defined as a \emph{leak}, is linear in \(k\): \(P(\text{leak}|k,\rho=0) = 1 - k/M\). It is obvious \(P(\text{leak}|k,0<\rho<1) < P(\text{leak}|k,\rho=0)\). This observation reveals two insights: (1) \(\alpha := P(\text{leak}|k,\rho)\) characterizes a type of risk in applying a top-\(k\) score filter to arms; (2) when controlling the risk of a leak at certain level, e.g.~\(\alpha=0.01\), \(k\) is a function of \(\rho\). The true correlation \(\rho\) between reward \(r\) and eligibility score \(e\) is unknown, but can be estimated from historical observations: \(\{(r_i,e_i)\}_i\). Therefore, EC essentially calculates \(k\) such that the risk \(P(\text{leak}|k,\hat{\rho})\) is controlled at a given level. 

To pair with K-Boot, EC also models \(P(\text{leak}|k,\rho)\) in a nonparametric fashion to avoid extra assumptions. Per \(\rho\), although Spearman's rank correlation coefficient is used for estimation, Kendall's \(\tau\) or any other rank-based correlation measure applies similarly. Across the arms and rounds, we assume the rank of \(e\) is a noisy perturbation of the rank of \(r\), and parameterize this perturbation as ``performing \(p\) random neighbor inversions''. An inversion is simply switching two elements in a sequence, and a neighbor inversion switches two neighboring elements. Note \(p = 0\) indicates \(\rho = 1\) because the perturbed rank sequence is identical to the original one; \(p \rightarrow \infty\) is effectively a random permutation thus \(\rho = 0\). This setup provides a generative process to simulate the joint distribution of all parameters for a rank sequence of length \(n\): (1) do \(p\) random neighbor inversions on the sorted sequence of ranks \((1,\cdots,n)\); (2) see if there is a \emph{leak}, namely if the element \(1\) is now at or beyond the \(k\)th position; (3) compute \(\hat{\rho}\) between the original and current sequences. The three steps can be replicated for a sufficient number of times, to obtain the final correlation coefficient \(\hat{\rho}\) and risk level \(\hat{\alpha}\) by averaging individual \(\hat{\rho}\) and counting the frequency of leaks. For different combinations of \((k,p,n)\), as the above process does not depend on any actual data, it can be performed offline in batch to get the corresponding \((\hat{\rho},\hat{\alpha})\). The resulting empirical dictionary \(G: (n,\alpha,\rho)\rightarrow k\) is stored, and later queried during online inference. To avoid unnecessary control in the edge case where eligibility score turns out to be pure noise, \(k\) is reset to \(M\) (no EC) if \(G\) outputs a \(k\) that is greater than the trivial value \((1-\alpha)M\). Finally note the generative distribution of random neighbor inversions is assumed to  be uniform along the whole sequence, yielding the unbiasedness of the  estimator \(\hat{\rho}\) even if the observed data points \(\{(r_i,e_i)\}_i\) are censored by the top-\(k\) filter policy.

\begin{algorithm*}[]
\DontPrintSemicolon
\SetAlgoLined
\SetKwInOut{Input}{Input}
\caption{Eligibility Control}
\Input{Number of iterations $N$, number of arms $M$, risk level $\alpha$, score initial threshold $k_0$, a pre-computed empirical dictionary $G(n,\alpha,\rho)\rightarrow k$}
\BlankLine
Initialize threshold $k := k_0$ and observation pool $\mathcal{D}_e := \varnothing$\;
\For{$n = 1,\dots,N$}{
    Observe context $\mathbf{x}_n$, and eligibility scores $\{e_1,\cdots,e_M\}$, for each arm\;
    Find the $k$th largest element $e_{[k]}$ among the scores\;
    Sample reward for each arm if $e_m \geq e_{[k]}$ (Algorithm 1, Line 5-18)\;
    Pull the $m^\star$th arm with highest reward estimate (Algorithm 1, Line 21-22)\;
    Observe true reward $r_n$, and update $\mathcal{D}_e := \mathcal{D}_e \cup \{(r_n,e_{m^\star})\}$\;
    (Periodically) compute $\hat{\rho}$ from $\mathcal{D}_e$ and set $k := G(n,\alpha,\hat{\rho})$\;
    Set $k := M$ if $k \geq (1-\alpha)M$\;
}
\end{algorithm*}

The full online EC process is described in Algorithm 2 (excluding the offline dictionary generation). The eligibility state interface is converted into the score interface before EC, so the scores are the only required inputs. The algorithm takes K-Boot (Algorithm 1) as the bandit counterpart just for demonstration, and applies to any bandit that has arm-independent reward models. EC has a single hyperparameter: the controlled risk of a leak \(\alpha\). Note $\alpha$ is the risk of missing \emph{the best} arm, so its influence on cumulative rewards depends on the arm reward distribution, specifically the gap between the first and second best arm. We leave other types of risk control mechanism with distributional assumptions to future work.


\section{Experiments}
\label {sec:Experiments}

In this section, we evaluate the performance of K-Boot with two benchmarks: LinUCB~\cite{li2010contextual} and NeuralUCB~\cite{zhou2019neuralucb}. The experiment setup mostly follows the methodology reported in \cite{zhou2019neuralucb} with 4 synthetic simulations and 4 realworld datasets. EC is tested on synthetic data with an eligibility score setup, then on an Amazon Customer Service routing dataset. Datasets are introduced below.

\textbf{Synthetic Datasets.} The bandit has $M = 10$ arms, and runs on $N = 5000$ samples/rounds. It observes context with $d = 20$ dimensions, each following independent $\mathcal{N}(0,1)$. Four types of true reward functions are tested - (1) \emph{linear}: $h_0(\mathbf{x}) = 0.1\mathbf{x}^T\mathbf{a}$; (2) \emph{quadratic}: $h_1(\mathbf{x}) = 0.05(\mathbf{x}^T\mathbf{a})^2$; (3) \emph{inner-product}: $h_2(\mathbf{x}) = 0.002\mathbf{x}^T\mathbf{A}^T\mathbf{A}\mathbf{x}$; (4) \emph{cosine}: $h_3(\mathbf{x}) = \text{cos}(3\mathbf{x}^T\mathbf{a}$); where each entry of the parameters $\mathbf{A} \in \mathbb{R}^{d\times d}$ and $\mathbf{a} \in \mathbb{R}^{d}$ is drawn from an independent $\mathcal{N}(0,1)$, different for each arm. The observed reward has noise: $r := h(\mathbf{x}) + \varepsilon$, where $\varepsilon \sim \mathcal{N}(0,\sigma_{r}^2)$ and $\sigma_{r}$ is drawn from $\mathcal{U}(0.01,0.5)$ independently for each arm. 20 runs are replicated by random seeds.

\textbf{Synthetic Datasets with Eligibility.} Given above, the eligibility scores are further simulated by perturbing the counter-factual reward of each arm: $e := wr + \varepsilon^\prime$, to induce the correlation. Here $w$ is an arm-specific scaling coefficient draw from $\mathcal{U}(-0.1,1)$. It has a small chance to leave $e$ and $r$ negatively correlated, as tolerating eligibility claim error in systems under the overall positive correlation - the only assumption of EC. The noise term $\varepsilon^\prime \sim \mathcal{N}(0,\sigma^{2}_{e})$ controls the correlation effect size. By setting a proper $\sigma_{e}$, the Spearman's rank correlation coefficient $\rho$ between reward and eligibility score can be fixed at an arbitrary positive value. 

\textbf{UCI Classification.} Four real-world datasets from UCI Machine Learning Repository \cite{dua2019uci} are gathered: \textit{covertype}, \textit{magic}, \textit{statlog}, and \textit{mnist}. These multi-class classification datasets are converted into multi-arm contextual bandits problems, following the method in \cite{li2010contextual}: each class is treated as an arm, and reward is $1$ if bandit pulls the correct arm (identify the correct class) and $0$ otherwise. Each dataset is shuffled and split into 10 mutually exclusive runs with 5000 samples each, and fed to bandit one in each round (\textit{magic} has less than 50000 samples so resampling is performed).

\begin{figure*}
     \centering
     \includegraphics[width=\textwidth]{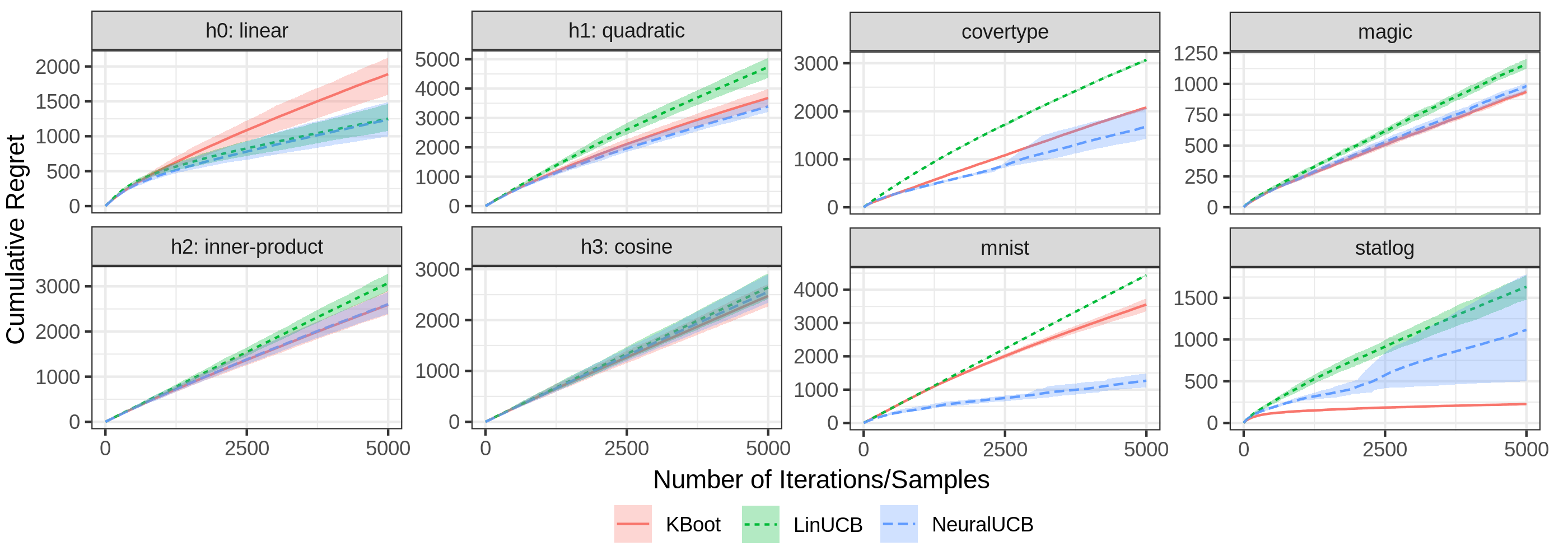}
     \caption{Benchmarking K-Boot, LinUCB and NeuralUCB. Left half: synthetic datasets; right half: UCI datasets. Model hyperparameters are set as the best on synthetic datasets. Metric is averaged across 10 runs with an 80\% confidence band.}
     \label{fig:figure2}
\end{figure*}

\textbf{CS Routing with Eligibility.} Agent skill-level routing is considered: CS agents are assigned to different skill groups by operation teams, and the routing system determines which skill is required to resolve a given customer contact based on the customer profile, policy rules and natural language understanding. As a proof of concept before any online experiments, we formulate an offline bandit dataset with historical observations. About 2 million Amazon US customer service text chats with bot-to-agent routing are gathered. To introduce stochasticity, a historical routing policy emulator is trained on a 95\% random subset of the chats, by taking the customer-bot transcripts before the routing action as input to predict which skill the system chose. Specifically, the emulator consists of a pretrained GPT-2 transformer encoder \cite{radford2019language}, with a multi-class sequence classifier head to predict the top 20 skill groups. The model is trained (fine-tuned) for one epoch with learning rate $10^{-4}$ and batch size 32. For the bandit simulation, the final actual resolving skill, after all the potential agent-to-agent transfers since the initial routing, serves as the ground truth - if the routed skill agrees with the final skill, the action is considered as accurate and having avoided a potential transfer. At each round, the bandit receives a 256-dimensional text embedding from the same GPT-2 encoder as observed context, and a 20-dimensional probabilistic prediction from the emulator as eligibility scores for each arm/skill. If the bandit chooses a skill matching the final one, reward is $1$ and otherwise $0$. In this case, the cumulative regret is an upper bound for the number of transferred contacts, because agents assigned to different skills may resolve the contact regardless (e.g. digital order agents are often equipped to resolve retail order issues, so they may not transfer an incorrectly routed contact). The rest 5\% chats are used to generate 10 bandit runs with 8000 samples each\footnote{From another perspective, this offline bandit-with-eligibility simulation setup is equivalent to "\textit{using a bandit for online improvement of a black-box probabilistic classifier with unknown accuracy"}.}. We observe that the empirical $\rho$ across the runs is 0.3492. We leave the below to future work: (1) online experiments to measure real transfers and positive customer reviews as the reward metric; (2) agent-level routing given agent profiles and finer grain eligibility definitions than skill groups.

\subsection{K-Boot Benchmarks}
\label{experiment1}

We compare K-Boot with LinUCB and NeuralUCB on both synthetic and UCI classification datasets. With the 4 synthetic datasets, we first select the following hyperparameters for each Bandit algorithm of interest (1) number of nearest neighbors $k \in \{20, 50, 100\}$ for K-Boot; (2) weight parameter for exploration $\alpha \in \{0.1, 1.0, 10\}$ for LinUCB; (3) regularization parameter $\lambda \in \{0.1, 0.01, 0.001\}$ and exploration parameter $\nu \in \{0.2, 0.02, 0.002\}$ for NeuralUCB - other hyperparameters such as learning rate, batch settings and number of layers are set the same as in the original code base\footnote{https://github.com/uclaml/NeuralUCB}. We settle on $k = 100, \alpha = 10, (\lambda, \nu) = (0.001, 0.002)$ per having the lowest cumulative regret averaged across $h_0\sim h_3$ and 10 replicated runs, and carry these values over to compare the actual model performance on the 4 UCI datasets. We believe this setup is more realistic because most bandit applications do not have the luxury for intensive hyperparameter tuning due to small sample size or non-stationary environment. Finally, all experiments in this paper implement online, single-sample model updates. Figure 2 left shows the performance of the models with final selected hyperparameters. Except for the linear reward scenario where LinUCB/NeuralUCB has advantage in correct/close model specification, there is no significant performance difference between K-Boot and NeuralUCB while LinUCB is inferior. Figure 2 right shows the performance comparison on the UCI real-world datasets. K-Boot and NeuralUCB have a tie of winning on two datasets each, and LinUCB is consistently the worst. By examing the 80\% confidence band (P10-P90 across 10 runs), we find NeuralUCB has larger performance volatility (wider bands) on real data than the other two models, due to learning instability in certain runs.

\subsection{Eligibility Control Testing}
\label{experiment2}

\begin{figure*}
     \centering
     \includegraphics[width=\textwidth]{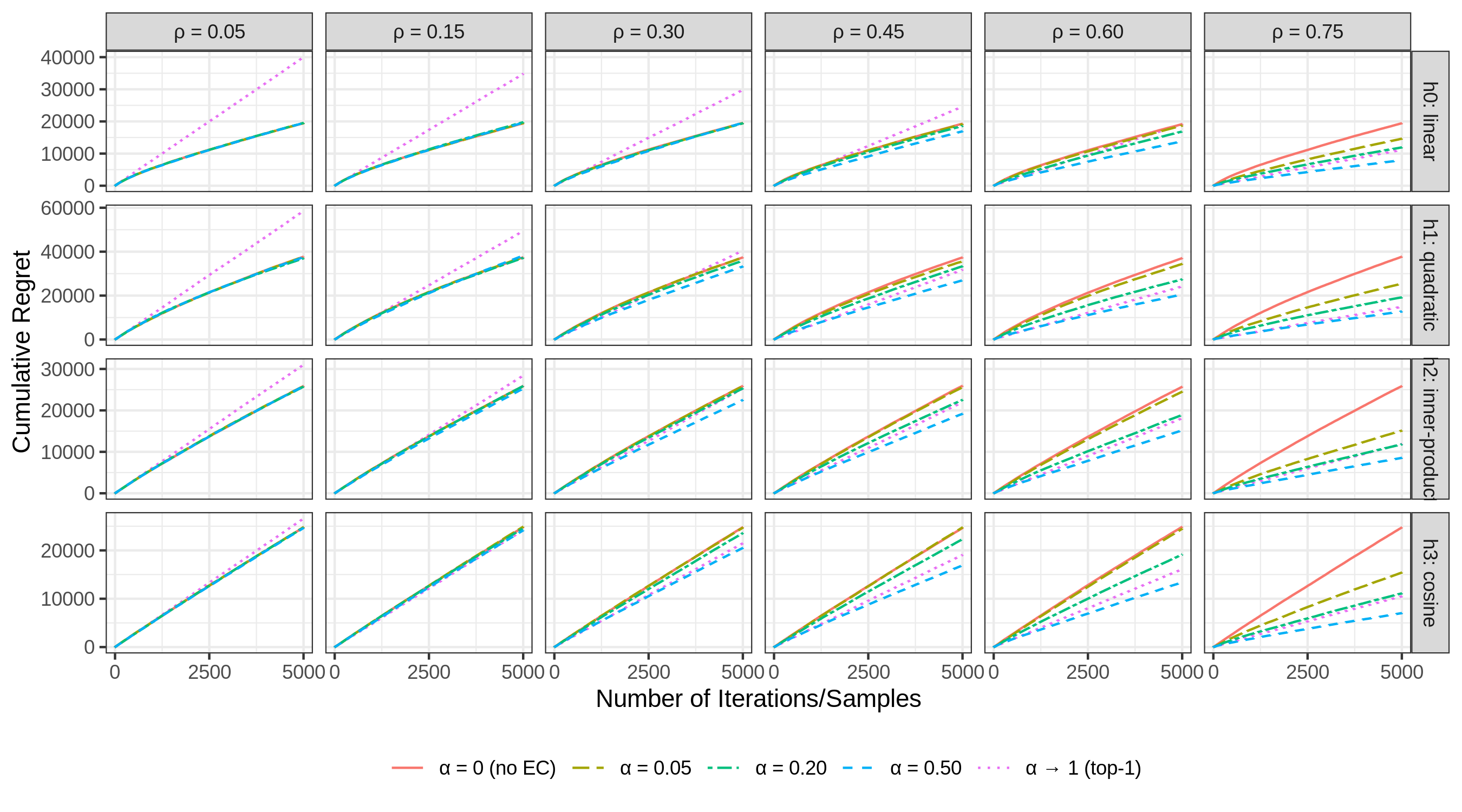}
     \caption{EC under different Spearman's Correlation $\rho$ between eligibility score and reward, with different risk level $\alpha$. The bandit part is K-Boot and eligibility scores are not in context.}
     \label{fig:figure3}
\end{figure*}

For the synthetic data, we control $\sigma_{e}$ to simulate different levels of reward-reflecting reliability in eligibility scores, with resulting $\rho(r,e) \in \{0.05,0.15,0.3,0.45,0.6,0.75\}$, for each true reward function $h_0$ - $h_3$. In the algorithm, we set different risk level $\alpha$ in EC. Note $\alpha = 0$ is effectively not using EC, and we abuse the notation $\alpha \to 1$ to denote the strict top-1 policy: pull the arm with the highest eligibility score. 

Figure 3 shows the results of applying EC to K-Boot, with eligibility scores used by EC only (explained later). When the signal in $e$ is weak (low $\rho$), EC adaptively raises the top-$k$ threshold close to $M$, achieving the same performance as no EC. As $\rho$ grows, the advantage reveals - using EC is consistently better than not. The dynamic thresholding may dominate the top-1 policy even under strong eligibility signals, with the right $\alpha$. EC performs the best for $\alpha = 0.5$ across all synthetic datasets, and the same value is carried over to the next experiment on CS Routing data.

EC can be applied to other bandits. Figure 4 shows the same synthetic data results but with EC + LinUCB. Here eligibility scores are also added to bandit context. The reason why the previous experiment did not take scores as K-Boot input is because $k$-NN has no native feature selection mechanism. If eligibility score is close to white noise (low $\rho$), model struggles to fit the reward, leading to a distraction from other presented visual patterns. For LinUCB, such side-effect is minimal for the simulated noise is Gaussian, so it is a better condition to test the difference between the trivial way versus the EC way of utilizing eligibility. EC still dominates no-EC in most cases. There are a few exceptions for the linear true reward function $h_0$, where $\alpha = 0.5$ is worse because LinUCB learns so well that taking a high risk of missing the best arm is not cost-effective. An interesting observation is that, while the strict top-1 rule was dominated by K-Boot + EC (Figure 3), it actually beats LinUCB + EC for nonlinear true reward functions even at $\rho = 0.15$. This indicates the power of the bandit model is still a key driver for ML to out-perform static rules.

\begin{figure*}
     \centering
     \includegraphics[width=\textwidth]{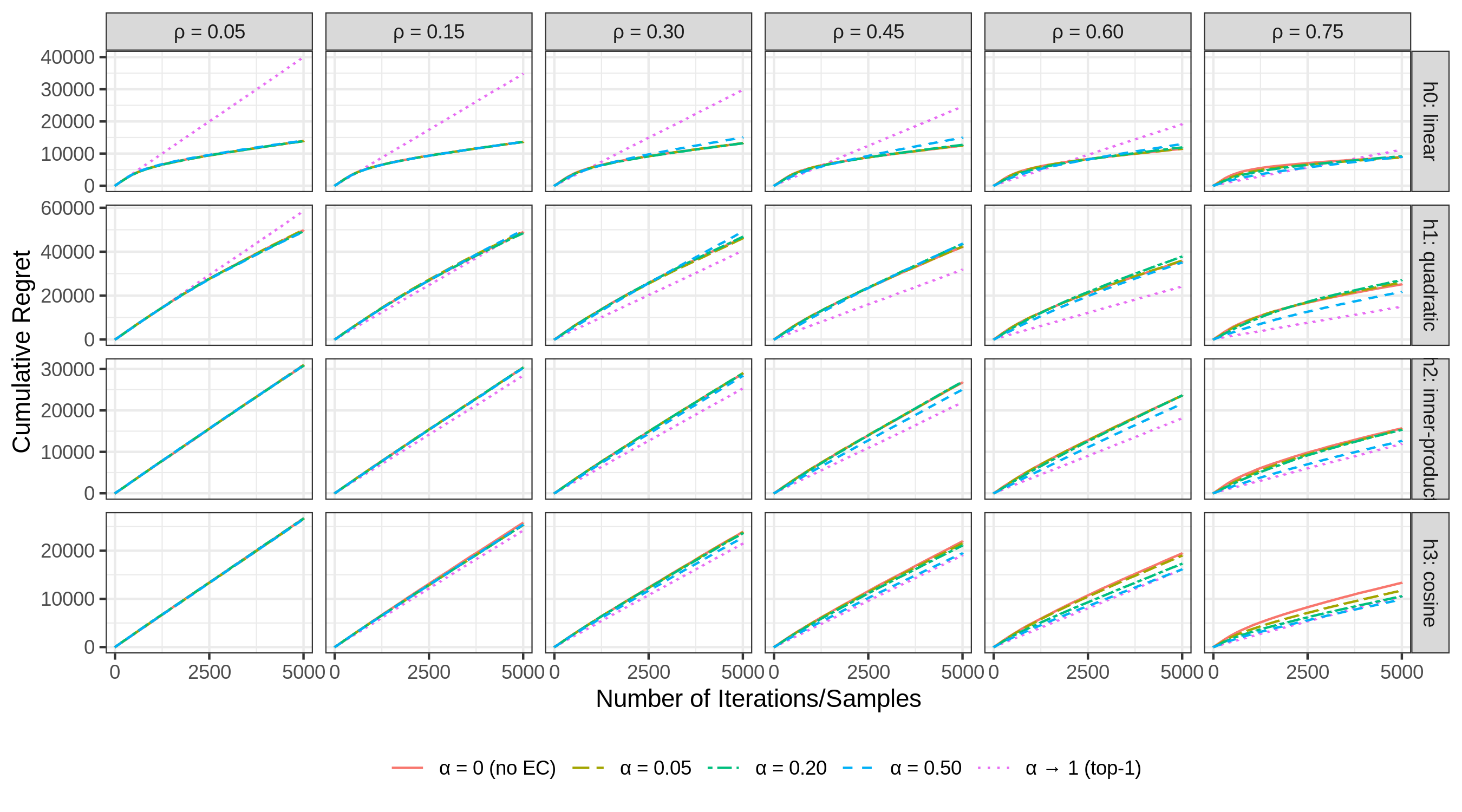}
     \caption{EC under different Spearman's Correlation $\rho$ between eligibility score and reward, with different risk level $\alpha$. The bandit part is LinUCB and eligibility scores are part of context.}
     \label{fig:figure4}
\end{figure*}

Figure 5 shows the result on CS Routing data, where EC significantly improves routing accuracy. Routing to the skill with the highest probability from the emulator serves a stochastic surrogate of the existing policy. Pure K-Boot result without knowing the emulator outputs is set as a baseline. With only 8000 samples, the average accuracy of the emulator policy is improved by K-Boot + EC from $53.37\%$ to $57.78\%$. 

\begin{figure}
     \centering
     \includegraphics[width=0.5\textwidth]{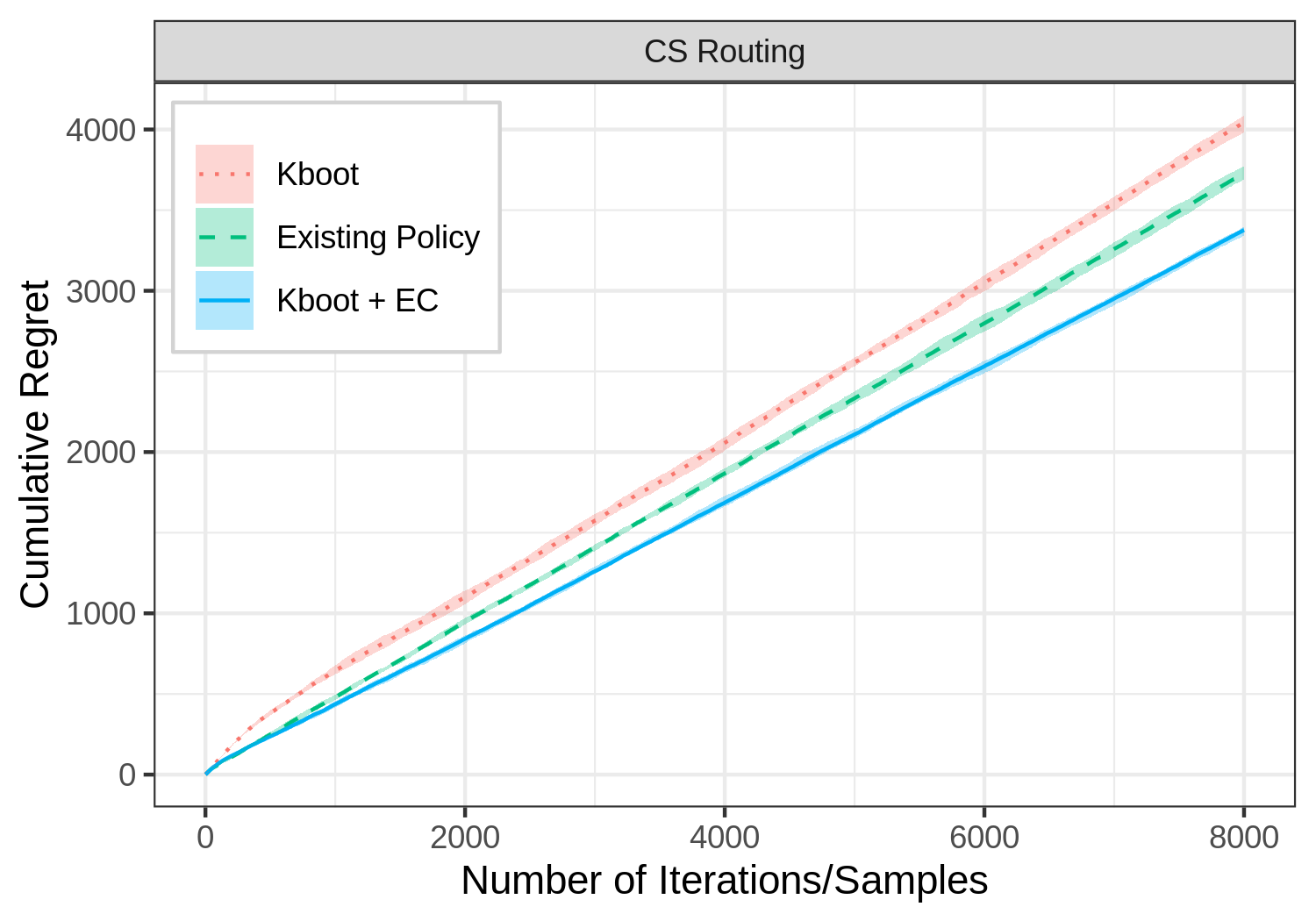}
     \caption{CS Routing Results. Metric is averaged across 10 runs with an 80\% confidence band.}
     \label{fig:figure5}
\end{figure}

\hypertarget{discussion-tbd}{%
\section{Conclusions}\label{discussion-tbd}}

We proposed a nonparametric contextual Bandit algorithm K-Boot with arm-level eligibility control (EC) for routing customer contacts to eligible SMEs in real-time. While K-Boot and EC are proposed here as a suite, each can be applied independently - K-Boot is a general Bandit algorithm and EC can control arm-eligibility for other bandits. We compared K-Boot with LinUCB and NeuralUCB. When looking at average regret performance over simulation runs, K-Boot and NeuralUCB had a tie in winning scenarios and were comparable in terms of robustness to different data distributions. Both worked better than LinUCB. However, we observe larger performance variability for NeuralUCB as opposed to K-Boot. We further found the EC component improved K-Boot's performance on both synthetic datasets and a simulation scenario in CS routing when eligibility exists.


\bibliography{bibliography}

\addcontentsline{toc}{section}{Reference}

\end{document}